\documentclass{article} 
\usepackage{iclr2027_conference,times}

\usepackage{amsmath,amsfonts,bm}

\def\eqref#1{equation~\ref{#1}}

\def\1{\bm{1}}

\DeclareMathAlphabet{\mathsfit}{\encodingdefault}{\sfdefault}{m}{sl}
\SetMathAlphabet{\mathsfit}{bold}{\encodingdefault}{\sfdefault}{bx}{n}

\usepackage{hyperref}
\usepackage{url}
\usepackage{booktabs}   
\usepackage{graphicx}   
\usepackage{multirow}
\usepackage{amssymb}

\title{AD-E2E-JEPA: A Joint-Embedding Predictive Architecture For End-to-End Autonomous Driving}

\author{
Haoran Zhu\textsuperscript{1},
Wancong Zhang\textsuperscript{1,2},
Yann LeCun\textsuperscript{1,2},
Anna Choromanska\textsuperscript{1}
\\
\textsuperscript{1}New York University
\qquad
\textsuperscript{2}AMI Labs
\\
\texttt{\{hz1922,wz1232,yann.lecun,ac5455\}@nyu.edu}
}

\iclrfinalcopy 
\begin{document}

\maketitle

\vspace{-0.1in}
\begin{abstract}
Autonomous driving requires \textit{world models} that can understand the physical world, reason and plan, and operate safely. In this paper, we first systematically evaluate existing action-conditioned joint-embedding predictive architecture (JEPA) world models, including LeWM, DINO-WM, and JEPA-WM for end-to-end autonomous driving (E2EAD). To isolate world-model quality from policy learning, we employ a goal-conditioned zero-shot planning setting that evaluates these models using ground-truth future observations as goals, without training any driving policy. We find that existing JEPA-based world models are either accurate for driving but computationally expensive, or computationally efficient but insufficient for planning. To address this trade-off, we propose \textbf{AD-E2E-JEPA}, which introduces a SIGReg-regularized learnable projector applied to projected patch embeddings. The projector reduces the number of planning patches by $16\times$ and the embedding dimension by $4\times$, achieving a $100\times$ inference speedup while retaining planning performance, with a 0.8-second runtime for an 8-frame rollout over 256 candidate trajectories. \textit{Without} training any driving policy, the world model itself reaches the goals located 20 meters away on average within the displacement of respectively 4.0/2.8 meters, using world-model rollouts over trajectory vocabularies of respectively 256/8,192 candidates. On the NAVSIMv2 benchmark, it achieves 67.3/72.9 EPDMS with multiplicative safety metrics and 84.1/86.5 EPDMS$^{\dagger}$ without them in goal-conditioned zero-shot planning. Experiments further show that the self-supervised pretrained projector improves downstream imitation learning performance from 80.2 to 85.4 EPDMS. The source code is available at \url{https://github.com/HaoranZhuExplorer/AD-E2E-JEPA}.
\end{abstract}

\vspace{-0.15in}
\begin{figure}[ht!]
  \centering
  \includegraphics[width=\linewidth]{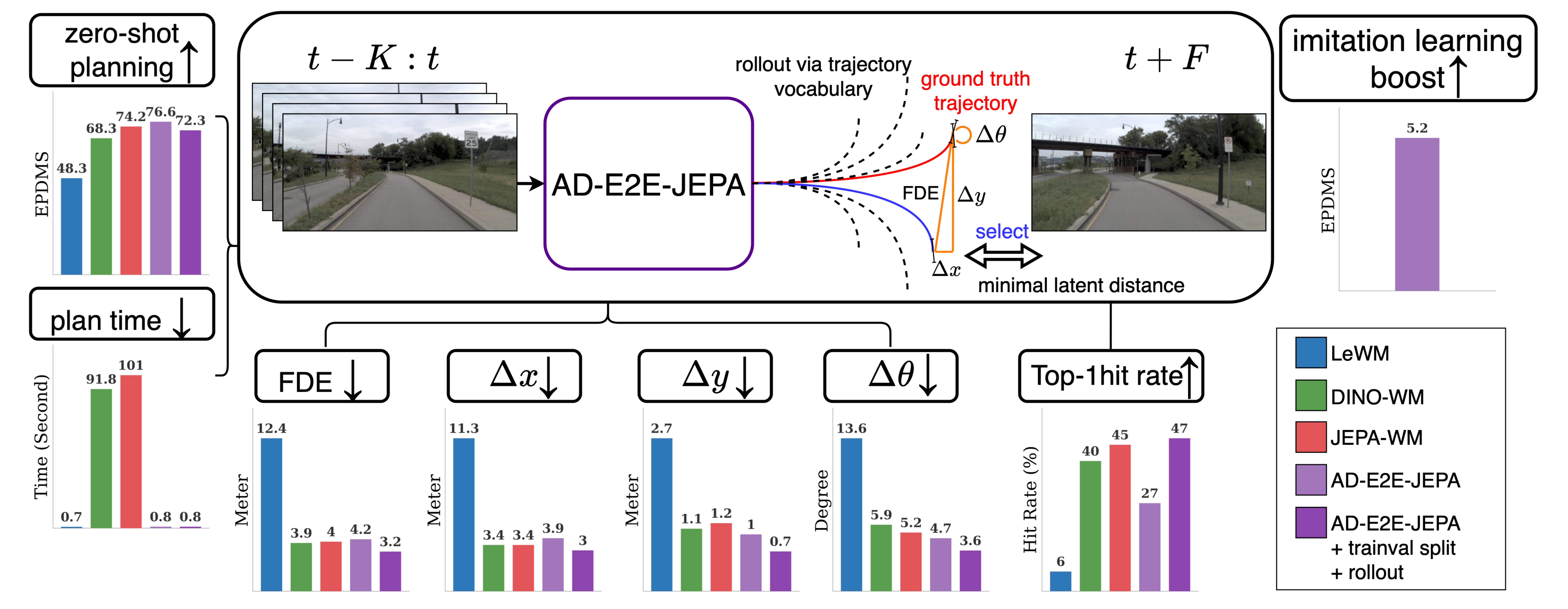}
  \caption{AD-E2E-JEPA is an action-conditioned, JEPA-based world model for end-to-end autonomous driving. \textit{Without} training any driving policy, it enables driving via goal-conditioned zero-shot planning, where future sensor observations are specified as the goal and a trajectory is selected from the trajectory vocabulary solely via world-model rollouts. AD-E2E-JEPA demonstrates strong zero-shot driving performance, efficient planning, low displacement errors, high trajectory hit rates, and representations that transfer effectively to downstream imitation learning. Zero-shot planning metrics are evaluated on 100 subsampled test scenes for a fair comparison with the time-consuming DINO-WM/JEPA-WM. See Table~\ref{tab:zero_shot_results} for AD-E2E-JEPA's results on the full set of 12,146 test scenes.}
  \vspace{-0.1in}
  \label{fig:overview}
\end{figure}

\pagebreak
\newpage

\section{Introduction}
Among intelligent systems operating in the physical world, autonomous vehicles, traditionally operated by humans, lie in the core interest of governments and the industry (Waymo, Tesla, NVIDIA, Wayve, XPENG, Uber, etc.). Efficient, robust, and safe autonomy defines the future of transportation leading to improved mobility of the population, reduced commute burdens, and increased road use. End-to-end learning~\citep{pomerleau1988alvinn, bojarski2016end, hu2023planning} has emerged as a popular paradigm for autonomous driving. Instead of relying on modular pipelines, an end-to-end autonomous driving (E2EAD) system~\citep{chen2024end} takes raw sensor observations as input and directly predicts a driving trajectory. However, existing E2EAD methods still rely heavily on imitation learning~\citep{hu2023planning, liao2025diffusiondrive}, where a reactive policy is trained to imitate human driving trajectories based on observed sensory context, without explicitly modeling the underlying dynamics of the driving environment. Moreover, human driving demonstrations can be suboptimal and noisy, which may limit the quality of supervision provided by imitation learning.

A promising direction beyond purely reactive imitation is to equip autonomous driving agents with a \textit{world model}~\citep{lecun2022path} based on a \textit{joint-embedding predictive architecture} (JEPA) that captures the underlying dynamics of the environment. Through self-supervised prediction in a latent embedding space, a driving agent can learn representations of the physical world by anticipating the future consequences of candidate driving trajectories or inferring environmental states that are not directly observable from sensor inputs. Such predictive objectives can yield representations useful for downstream tasks. More importantly, an action-conditioned world model can be rolled out to predict possible future states under different candidate actions, enabling the agent to evaluate alternative trajectories and select actions that best achieve a desired goal. This provides a principled foundation for planning and decision-making beyond direct imitation of human demonstrations.

In this paper, we present AD-E2E-JEPA (\textbf{A}utonomous \textbf{D}riving with an \textbf{E}nd-\textbf{to}-\textbf{E}nd \textbf{J}oint-\textbf{E}mbedding \textbf{P}redictive \textbf{A}rchitecture), an action-conditioned JEPA-based world model for autonomous driving. Unlike imitation-learning-based reactive systems, AD-E2E-JEPA uses human driving trajectories only as action conditioning for world-model learning, rather than as supervision for training a driving policy. Without training an explicit driving policy, the learned world model enables goal-conditioned zero-shot planning by selecting among candidate driving trajectories to navigate toward a goal specified by an image. Compared with existing JEPA world models, AD-E2E-JEPA introduces a learnable projector with SIGReg regularization, which substantially reduces planning latency while preserving planning performance. Moreover, the self-supervised pretrained projector transfers effectively to downstream imitation-learning-based E2E driving.

Our contributions can be summarized as follows:
\begin{itemize}

\item We are the first to adapt JEPA to E2EAD for goal-conditioned zero-shot planning. Beyond the success rate commonly used for JEPA-based world models, we also report driving performance metrics such as planning efficiency, geodesic accuracy, and reliability.
\item We show that existing JEPA-based world models face a substantial trade-off between efficiency and performance for planning: they are either computationally efficient but inaccurate at planning, or accurate but require over a minute of planning per scene.
\item To address this trade-off, we propose AD-E2E-JEPA, which builds on~\citet{wang2026temporal} by reducing embedding dimensionality and further reducing the number of patch embeddings while leveraging SIGReg regularization~\citep{balestriero2025lejepa} to preserve planning performance. This design substantially improves planning efficiency and may be applied to other JEPA-based world-model planning domains beyond E2EAD.
\item For goal-conditioned zero-shot planning on the NAVSIM test set, AD-E2E-JEPA significantly improves planning performance over LeWM, a gain of 23.7 EPDMS averaged over all 12,146 testing scenes, while requiring $100\times$ less planning time over DINO-WM/JEPA-WM. Our best variant achieves a 72.9 EPDMS, 2.8-meter average displacement error, and a 2.0-degree mean absolute heading error when selecting from 8192 candidate trajectories.
\item We further show that the self-supervised pretrained projector transfers effectively to downstream imitation learning. Integrated into a simple ViT-based E2E architecture, it improves EPDMS from 80.2 to 85.4 compared with a random projector, demonstrating that world-model representations can also benefit imitation-learning-based E2E driving.
\end{itemize}

\section{Related Work}
\subsection{World Models}

World models are internal representations that enable an agent to predict what is likely, plausible, or impossible, thereby providing a foundation for what is often referred to as common sense~\citep{lecun2022path}. The idea of world models can be traced back to a long history of planning and control~\citep{bryson1975applied, sutton1991dyna}. Early world models performed predictions directly in pixel space for robotic planning~\citep{finn2017deep} or learned latent dynamics using pixel reconstruction objectives~\citep{ha2018world}. More recent work suggests that learning world models purely in latent space, without reconstructing pixels, can lead to improved planning performance~\citep{zhou2024dino}. Generative world models have also recently received considerable attention. For example, current generative world models employ diffusion transformers~\citep{peebles2023scalable} to generate high-fidelity videos. However, despite their visual realism, whether such generative models reliably capture physical laws and can consequently benefit planning remains unclear~\citep{kang2024far}.

Joint-Embedding Predictive Architecture (JEPA)~\citep{lecun2022path} enables world models to be learned directly in latent space by predicting the representation of target data $y$ from context data $x$, with regularization-based methods~\citep{bardes2021vicreg,balestriero2025lejepa,kuang2026rectified,wu2026visreg} maximizing information to avoid representation collapse, in which all representations converge to a constant vector and become useless. Recent studies suggest that intuitive physics can emerge from JEPA~\citep{garrido2025intuitive}. JEPA-based approaches have been explored across multiple modalities, including images~\citep{assran2023self}, videos~\citep{bardes2024revisiting,assran2025v,mur2026v}, LiDAR~\citep{zhu2026self}, and audio~\citep{fei2023jepa,wang2026music}. These latent representations have further been shown to support zero-shot planning~\citep{zhou2024dino,sobal2026learning,maes2026leworldmodel} with action-conditioned world models, in which the model predicts future latents conditioned on actions. Recent work has investigated additional structural properties of the latent space that can facilitate planning. Temporal straightening~\citep{wang2026temporal}, for example, regularizes latent trajectories toward straighter paths, and prior work shows that projecting embeddings into a lower-dimensional space can improve planning performance. Other recent studies have demonstrated the potential of hierarchical planning~\citep{zhang2026hierarchical}, the benefits of sparse representations for planning~\citep{kuang2026lpwm}, and the benefits of adaptively updating world models during planning~\citep{wang2026adajepa}.

\subsection{End-to-End Autonomous Driving}

End-to-end autonomous driving (E2EAD) directly maps raw sensor inputs to planning actions using a fully differentiable model~\citep{chen2024end}, reducing error accumulation across modules and enabling joint optimization for planning~\citep{hu2023planning}. Most existing methods rely heavily on imitation learning to mimic human driving policies~\citep{chitta2022transfuser,hu2023planning,jiang2023vad,chen2024vadv2}. More recently, scoring-based methods rank candidate trajectories using annotated driving scores~\citep{li2024hydra,li2025hydra,kirby2026driving}, but still need imitation learning and dense trajectory-level supervision. World models have also gained attention in E2EAD, mainly for aligning predicted and ground-truth future representations~\citep{li2024enhancing,li2024navigation} or generating future driving videos~\citep{zhang2025epona,hu2023gaia}.

JEPA has recently been explored for autonomous driving. AD-L-JEPA~\citep{zhu2026self} firstly introduces JEPA pre-training for LiDAR perception, then Drive-JEPA~\citep{wang2026drive,naeinian2026zero} applies JEPA-based representations to end-to-end driving. AD-LiST-JEPA~\citep{zhu2026adlistjepa} extends JEPA to temporal LiDAR world modeling, while Auto-JEPA~\citep{yang2026auto} and DA-WAM~\citep{zhong2026wam} predict future representations for scoring-based planning. WA-JEPA~\citep{wang2026wa} combines representation learning, future prediction, and imitation learning. In contrast, our work directly investigates the zero-shot planning capability of JEPA without task-specific imitation learning.

\section{Method}
\begin{figure}[ht!]
  \centering
  \includegraphics[width=\linewidth]{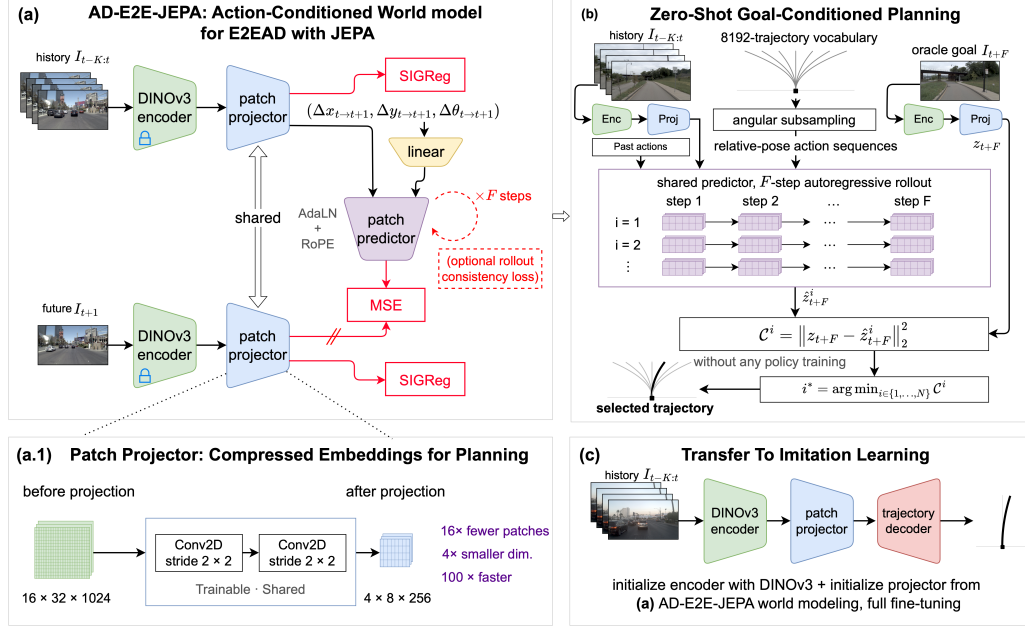}
  \caption{AD-E2E-JEPA architecture: \textbf{(a)} A self-supervised, action-conditioned JEPA world model for E2EAD. The DINOv3 encoder is frozen, while \textbf{(a.1)} a learnable projector is shared by the context-history and target-future branches, compressing the representation by $16\times$ and its dimensionality by $4\times$, with SIGReg maximizing information while avoiding collapse and improving planning performance. This enables a $100\times$ speedup in zero-shot goal-conditioned planning in \textbf{(b)}, where trajectories are selected via world-model rollouts over a driving vocabulary. \textbf{(c)} The projector pretrained through self-supervision also improves downstream imitation learning.}
  \label{fig:architecture}
\end{figure}

We propose AD-E2E-JEPA, illustrated in Figure~\ref{fig:architecture}, a joint-embedding predictive architecture (JEPA) for end-to-end autonomous driving (E2EAD). It enables efficient zero-shot goal-conditioned planning, while its self-supervised pretrained projector also benefits downstream imitation learning.

\subsection{Preliminary: Task formulation of E2EAD}
For simplicity, we consider a front-camera-only setting. At each time step \(t\),
the driving agent observes a context window of historical image-pose pairs and
predicts a sequence of future ego poses that form a driving trajectory:
\begin{align}
    \text{Historical context:} \quad
    &(\mathbf{I}_{t-K:t},\, \mathbf{P}_{t-K:t}) \\
    \text{Trajectory prediction:} \quad
    &\hat{\mathbf{P}}_{t+1:t+F}
    = f_\phi(\mathbf{I}_{t-K:t},\, \mathbf{P}_{t-K:t})
\end{align}
Here, \(\mathbf{I}_j \in \mathbb{R}^{3\times H\times W}\) denotes the
front-camera image at time step \(j\), and
\(\mathbf{P}_j=[x_j,y_j,\theta_j]^\top\in\mathbb{R}^3\) denotes the
corresponding ego pose, consisting of the planar position \((x_j,y_j)\) and
heading angle \(\theta_j\) in radians. All poses are expressed relative to the
current ego pose at time step \(t\), such that
\(\mathbf{P}_t=[0,0,0]^\top\). The objective of E2EAD is to
predict future ego poses that safely make progress.

\subsection{World Model Architecture}
\subsubsection{JEPA-WM Adaptation for E2EAD}
We initialize our world model baseline using the best-performing configuration identified in JEPA-WM~\citep{terver2025drives}, which conducts extensive ablations on the design choices of JEPA-based world models and finds that the optimal configuration uses DINOv3~\citep{simeoni2025dinov3} ViT-L as the backbone encoder, together with an AdaLN-style~\citep{perez2018film,peebles2023scalable} predictor equipped with RoPE~\citep{su2024roformer}, optionally with rollout training. DINOv3 is preferred over DINOv2~\citep{oquab2023dinov2} and V-JEPA 2~\citep{assran2025v}, potentially due to its dense semantic representations, which are also well suited to autonomous driving in our setting.

To adapt the model to E2EAD, we define the action at each time step as the relative pose change between consecutive frames, following~\citet{zhang2025epona}, as shown in Eq.~(\ref{eq:e2ead_action}):
\begin{equation}
\label{eq:e2ead_action}
\mathbf{a}_t
=
\operatorname{Relative}(\mathbf{P}_t,\mathbf{P}_{t+1})
=
\left[
\Delta x_{t \rightarrow t+1},
\Delta y_{t \rightarrow t+1},
\Delta \theta_{t \rightarrow t+1}
\right]^{\top}.
\end{equation}

The frozen DINOv3 backbone independently encodes the history frames and the subsequent frame, as shown in Eq.~(\ref{eq:encoding}). Given the history embeddings and action embeddings produced by a linear layer $E_a$, the AdaLN predictor predicts the embedding sequence one time step ahead, as shown in Eq.~(\ref{eq:prediction}). We supervise the predictions using only an MSE loss, as defined in Eq.~(\ref{eq:prediction_loss}), without an additional anti-collapse objective, since the prediction targets are from the frozen DINOv3 encoder:
\begin{align}
s_{t-K:t+1}
&= \operatorname{Enc}(\mathbf{I}_{t-K:t+1}),
\label{eq:encoding} \\
\hat{s}_{t-K+1:t+1}
&= \operatorname{Pred}\!\left(
s_{t-K:t},
E_a(\mathbf{a}_{t-K:t})
\right),
\label{eq:prediction} \\
\mathcal{L}_{\mathrm{pred}}
&= \operatorname{MSE}\!\left(
\hat{s}_{t-K+1:t+1},
s_{t-K+1:t+1}
\right).
\label{eq:prediction_loss}
\end{align}

\subsubsection{AD-E2E-JEPA}
We propose AD-E2E-JEPA for efficient and reliable planning. Existing JEPA-based world model baselines operate on dense patch embeddings produced by the encoder, making planning computationally expensive and potentially requiring several minutes~\citep{assran2025v}. This is impractical for the E2EAD setting and motivates embedding compression. Experiments in~\citet{wang2026temporal} show that adding a projector with two convolutional layers with stride $1\times1$ can reduce embedding dimensionality while improving planning performance. Inspired by this, AD-E2E-JEPA further introduces a learnable projector $\operatorname{Proj}(\cdot)$ consisting of two convolutional layers with stride $2\times2$. The projector is applied independently to the encoder embeddings of the history and future frames, as shown in Eq.~(\ref{eq:projection}), reducing their spatial size by $16\times$ while also reducing the ViT-L embedding dimensionality by $4\times$, from $D_{\text{enc}}=1024$ to $D_{\text{proj}}=256$.

\vspace{-0.2in}
\begin{equation}
    \label{eq:projection}
    z_{t-K:t+1}
    =
    \operatorname{Proj}(s_{t-K:t+1})
\end{equation}
\vspace{-0.2in}

The predictor operates on the projected embeddings, as shown in Eq.~(\ref{eq:rollout_proj}). 

\vspace{-0.2in}
\begin{equation}
    \label{eq:rollout_proj}
    \hat{z}_{t-K+1:t+1}
    =
    \operatorname{Pred}^{\prime}\!\left(
    z_{t-K:t},
    E_a(\mathbf{a}_{t-K:t})
    \right)
\end{equation}
\vspace{-0.2in}

The prediction loss is then defined over the projected embeddings in Eq.~(\ref{eq:loss_pred_proj}). However, in this case, the projected embeddings may suffer from representation collapse, where distinct DINOv3 embeddings are mapped to nearly constant vectors and thus become uninformative. We apply stop gradient~\citep{chen2021exploring, grill2020bootstrap} to the projected target embedding in the MSE loss.

\begin{equation}
    \label{eq:loss_pred_proj}
    \mathcal{L}^{\mathrm{proj}}_{\mathrm{pred}}
    =
    \operatorname{MSE}\!\left(
    \hat{z}_{t-K+1:t+1},
    \operatorname{sg}\left(z_{t-K+1:t+1}\right)
    \right)
\end{equation}

Furthermore, we apply SIGReg~\citep{balestriero2025lejepa} to encourage the embeddings to follow an isotropic Gaussian distribution and prevent representation collapse. Unlike the SIGReg formulation in~\citet{maes2026leworldmodel}, which is applied to global CLS embeddings across the batch independently at each time step and then averaged over time, we apply SIGReg to patch embeddings independently at each patch location and time step across the batch, and average the resulting regularization over all time steps and patch locations, as shown in Eq.~(\ref{eq:sigreg}):

\vspace{-0.2in}
\begin{equation}
\label{eq:sigreg}
\mathcal{L}^{t-K:t+1}_{\mathrm{SIGReg}}
=
\frac{1}{N H^{\prime} W^{\prime} M}
\sum_{l=1}^{N H^{\prime} W^{\prime}}
\sum_{m=1}^{M}
T\left(
\left\{
\left\langle z_{l,b}, \boldsymbol{u}^{(m)} \right\rangle
\right\}_{b=1}^{B}
\right).
\end{equation}
\vspace{-0.15in}

For the single-step setting in Eq.~(10), $l$ indexes the projected patch locations across the temporal window of length $N = K + 2$ and spatial dimensions $H' \times W'$. $T(\cdot)$ denotes the univariate Epps--Pulley test~\citep{epps1983test}, which regularizes the $D$-dimensional embeddings across the batch of size $B$ using $M$ random projection directions $\boldsymbol{u}^{(m)} \in \mathbb{S}^{D-1}$. By the Cram\'er--Wold theorem~\citep{cramer1936some}, matching all one-dimensional projected distributions is equivalent to matching the full joint distribution; SIGReg approximates this objective using a finite number of random projections. A detailed configuration of SIGReg is provided in Appendix~\ref{appendix:sigreg}.

The overall training loss for AD-E2E-JEPA is given by Eq.~(\ref{eq:overall_loss}):
\begin{equation}
\label{eq:overall_loss}
\mathcal{L}
=
\mathcal{L}^{\mathrm{proj}}_{\mathrm{pred}}
+
\lambda\mathcal{L}^{t-K:t+1}_{\mathrm{SIGReg}}.
\end{equation}

We optionally add rollout training, as in JEPA-WM~\citep{terver2025drives}, over the entire future horizon $F$ for E2EAD. The training objective consists of a teacher-forcing prediction loss, $\overline{\mathcal{L}}_{\mathrm{TF}}$, over the full future horizon from the next frame through frame $t+F$. We also apply rollout losses, $\overline{\mathcal{L}}_{2}+\cdots+\overline{\mathcal{L}}_{F}$, based on autoregressive predictions from frame $t+2$ through frame $t+F$. These rollout losses encourage consistency. In addition, we apply SIGReg from frame $t-K$ through frame $t+F$. The overall loss is given by Eq.~(\ref{eq:overall_loss_with_rollout}), with further details provided in Appendix~\ref{appendix:loss_with_rollout}.

\vspace{-0.2in}
\begin{equation}
\label{eq:overall_loss_with_rollout}
\mathcal{L}^{\mathrm{multi}}
=
\frac{
\overline{\mathcal{L}}_{\mathrm{TF}}
+
\overline{\mathcal{L}}_{2}
+
\cdots
+
\overline{\mathcal{L}}_{F}
}{F}
+
\lambda \mathcal{L}^{t-K:t+F}_{\mathrm{SIGReg}}.
\end{equation}
\vspace{-0.2in}

\subsection{Zero-Shot Goal-Conditioned Planning}
World models enable zero-shot goal-conditioned planning \textit{without} training a policy, allowing generalization to unseen scenes. This paradigm has been widely adopted in JEPA-based world models. We are the first to introduce it to E2EAD, significantly improve planning efficiency, and report several quantitative metrics to evaluate world-model beyond the success-rate metric used in prior work.

\subsubsection{Planning with AD-E2E-JEPA}
\label{sec:planning}
The cross-entropy method (CEM) is widely used for JEPA-based world-model planning~\citep{zhou2024dino, terver2025drives, sobal2026learning} but is too costly for autonomous driving due to its iterative evaluation of many candidate action sequences. We instead search over the clustered driving trajectory vocabulary as anchors from~\citet{chen2024vadv2},
$\mathcal{V}=\{P^i_{t:t+F}\}_{i=1}^{8192}$.
To balance efficiency and trajectory granularity, we sort trajectories by angular coordinate and subsample them at evenly spaced intervals to obtain e.g., 
$|V_\text{sampled}|=256$ candidates.

We define the goal as the future image $I_{t+F}$, $F$ frames ahead. Given our action-conditioned world model, we perform zero-shot planning by selecting the candidate whose predicted future latent embedding is closest to that of the goal. For the $i$-th candidate trajectory, the planning cost is

\vspace{-0.2in}
\begin{equation}
    \label{eq:planning_cost}
    \mathcal{C}^{i}
    =
    \left\| z_{t+F} - \hat{z}_{t+F}^{\,i} \right\|_2^2,
\end{equation}
where $\hat{z}_{t+F}^{\,i}$ is obtained by autoregressively rolling out the world model from the initial projected observations $z_{t-K:t}$ under the $i$-th candidate trajectory using Eq.~(\ref{eq:rollout_proj}). We select the trajectory as

\vspace{-0.2in}
\begin{equation}
    i^{*}
    =
    \operatorname*{argmin}_{i \in \{1,\ldots,|V_\text{sampled}|\}}
    \mathcal{C}^{i}
    =
    \operatorname*{argmin}_{i \in \{1,\ldots,|V_\text{sampled}|\}}
    \left\| z_{t+F} - \hat{z}_{t+F}^{\,i} \right\|_2^2,
    \qquad
    P^{*}_{t:t+F} = P^{i^{*}}_{t:t+F}.
\end{equation}
\vspace{-0.1in}

\subsubsection{Metrics}
\label{sec:metrics}

To quantitatively evaluate the reliability and efficiency of world-model zero-shot goal-conditioned planning for autonomous driving, we report metrics for driving performance (EPDMS, EPDMS$^\dagger$), planning efficiency, geodesic accuracy. (FDE, $\Delta x$, $\Delta y$, $\Delta \theta$), and reliability (hit rate). Detailed definitions are provided in Appendix~\ref{appendix:metrics}.

\textbf{EPDMS}: NAVSIMv2~\citep{cao2025pseudo} uses EPDMS as a pseudo-simulation-based planning metric that combines multiplicative safety terms with a weighted measure of driving quality.

\textbf{EPDMS$^\dagger$}: Since zero-shot planning does not explicitly optimize for safety, we report only the weighted component of EPDMS, excluding the multiplicative safety terms.

\textbf{Planning time}: We report planning time to assess E2EAD planning efficiency.

\textbf{Final-pose displacement (FDE, $\Delta x$, $\Delta y$, $\Delta \theta$)}:
We measure the displacement between the final pose of the selected trajectory and the ground-truth pose. Specifically, we report the final displacement error (FDE) and absolute errors in longitudinal position ($\Delta x$), lateral position ($\Delta y$), and heading ($\Delta \theta$), which measure the geodesic accuracy. of the world model.

\textbf{Hit rate}:
Inspired by the diagnostic metric released in the DrivoR~\citep{kirby2026driving} codebase, we use hit rate to measure whether the world model ranks the ground-truth trajectory among the top-$k$ lowest-cost candidates. Specifically, we report Top-1 and Top-5 hit rates based on the rollout latent distance to the goal, which assess the reliability of the world-model rollout.

\subsection{Downstream Transfer for Imitation Learning}
Beyond zero-shot goal-conditioned planning, we evaluate whether the self-supervised projector pretrained during world modeling provides useful representations for imitation learning. Existing works~\citep{wang2026drive, naeinian2026zero} show that self-supervised pretrained representations improve E2EAD, but have not explored projectors that compress dense patch embeddings. 

We discard the patch predictor from AD-E2E-JEPA while retaining the DINOv3 encoder and projector, attach a simple trajectory decoder, and fully fine-tune the model for E2EAD. We compare against variants with a randomly initialized projector.

\section{Experiments}

\subsection{Datasets}
We use the NAVSIM~\citep{cao2025pseudo} dataset and evaluate on the latest NAVSIMv2 benchmark. We use the \texttt{navtrain} split, which contains 10 hours of driving video sampled at 2~Hz, for a fair comparison among LeWM, DINO-WM, JEPA-WM, and AD-E2E-JEPA, while scaling AD-E2E-JEPA to 70 hours of driving video from the training portion of the \texttt{trainval} split to further improve its performance.

\subsection{Implementation Details}
We use DINOv3 ViT-L backbone for DINO, JEPA-WM and AD-E2E-JEPA. LeWM's backbone is ViT-L trained from scratch. We use $K+1=4$ frames for a 2-s history context and $F=8$ frames for a 4-s future horizon. Without the optional rollout loss, AD-E2E-JEPA uses only the first future frame in Eq.~(\ref{eq:overall_loss}); the rollout-loss variant uses all 8 future frames in Eq.~(\ref{eq:overall_loss_with_rollout}). We use AdamW with 30 training epochs for all settings and scale the learning rate with the square root of the batch size. We use one warmup epoch followed by a cosine annealing schedule. The default SIGReg weight is $\lambda=0.09$; for larger batch sizes, we tune $\lambda$ heuristically based on early training loss curves. For LeWM, DINO-WM, and JEPA-WM, we use a learning rate of $1\times10^{-4}$ across all settings with the same learning-rate scheduler. AD-E2E-JEPA requires fewer GPU resources hours than the other methods when training settings are the same. Table~\ref{tab:training_details} summarizes the training configurations. For transferring self-supervised pretrained projectors to imitation learning, see Appendix~\ref{appendix:imitation_learning}.

\begin{table}[t]
    \centering
    \caption{Training configurations.}
    \label{tab:training_details}
    \begin{tabular}{llccccc}
        \toprule
        Split & Variant & GPUs & Batch size & Learning rate & $\lambda$ & Training time \\
        \midrule

        \multirow{5}{*}{\texttt{navtrain}}
        & LeWM
        & $4\times$A100
        & 8
        & $1\times10^{-4}$
        & 0.09
        & 1 d \\

        & DINO-WM
        & $4\times$A100
        & 64
        & $1\times10^{-4}$
        & --
        & 11 h \\

        & JEPA-WM
        & $4\times$A100
        & 64
        & $1\times10^{-4}$
        & --
        & 13 h \\

        & AD-E2E-JEPA
        & $1\times$A100
        & 128
        & $1\times10^{-4}$
        & 0.09
        & 20 h \\

        & \quad\quad + rollout
        & $1\times$A100
        & 128
        & $1\times10^{-4}$
        & 0.09
        & 1 d 22 h \\

        \midrule

        \multirow{2}{*}{\texttt{trainval}}
        & AD-E2E-JEPA
        & $4\times$A100
        & 512
        & $2\times10^{-4}$
        & 0.025
        & 2 d 5 h \\

        & \quad\quad + rollout
        & $4\times$A100
        & 256
        & $1.4\times10^{-4}$
        & 0.09
        & 4 d 2 h \\

        \bottomrule
    \end{tabular}
    \vspace{-0.15in}
\end{table}

\subsection{Zero-Shot Planning Performance}

\begin{figure}[ht!]
  \centering
  \includegraphics[width=\linewidth]{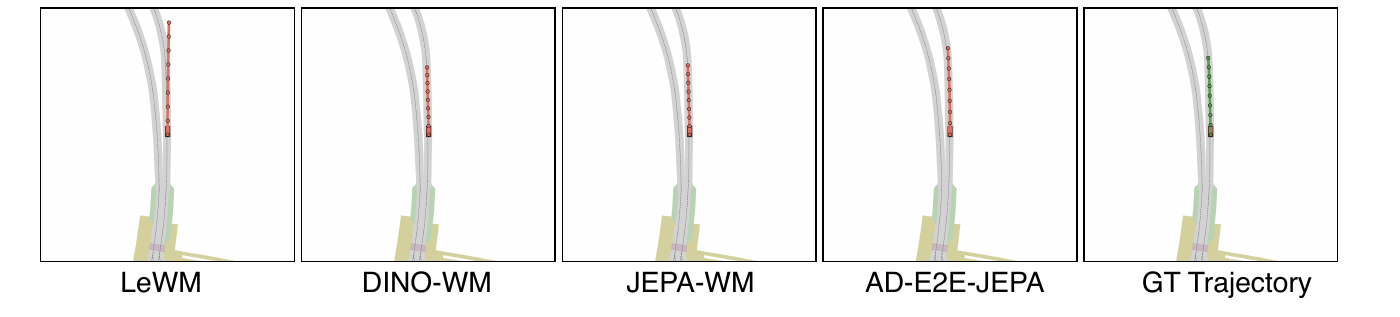}
  
  \caption{Zero-shot goal-conditioned planning qualitative comparison. LeWM is efficient but insufficient for accurate planning, resulting in a shifted selected trajectory. DINO-WM and JEPA-WM better match the ground-truth trajectory. AD-E2E-JEPA achieves comparable planning quality to DINO-WM and JEPA-WM while being 100$\times$ faster.}
  \label{fig:visualization}
\end{figure}

We set the number of subsampled trajectories to 256 unless specified. We evaluate zero-shot goal-conditioned planning on 100 sampled scenes from the NAVSIM test split for all methods. We additionally evaluate the computationally efficient models on the full test set. DINO-WM and JEPA-WM operate on dense patch embeddings after the encoder, making full-set evaluation prohibitively expensive (over 10 days). AD-E2E-JEPA performs planning following Section~\ref{sec:planning}. LeWM, DINO-WM, and JEPA-WM follow the same planning procedure, except that DINO-WM and JEPA-WM operate on patch embeddings, whereas LeWM operates on the global CLS token. We omit these formulations for brevity. The qualitative planning visualization across these methods is in Figure~\ref{fig:visualization}.

\begin{table*}[ht!]
\caption{Zero-shot goal-conditioned planning on the NAVSIM dataset, evaluated on 100 subsampled test scenes and the full set of 12,146 test scenes, with the ground-truth future frame used as the target. Unless specified, the world model selects from 256 candidate trajectories. The NAVSIM-v2 \texttt{navtest-stage-1} metrics EPDMS$\uparrow$ and EPDMS$^\dagger\uparrow$ measure driving performance, where EPDMS includes safety metrics and EPDMS$^\dagger$ excludes them. Per-scene planning time (s) measures efficiency on an  A100 GPU. FDE (m), $\Delta x$ (m), $\Delta y$ (m), and $\Delta \theta$ (degrees) measure geodesic accuracy. Top-1/top-5 hit rate (\%) measures world-model reliability. $\uparrow$: higher is better; $\downarrow$: lower is better. For the 100 subsampled-test scene setting, the extended comfort (EC) metrics for EPDMS and EPDMS$^\dagger$ are excluded because the subsampled set does not contain temporally adjacent scenes. See Appendix~\ref{appendix:epdms_details} for details of each submetric of EPDMS.}
\label{tab:zero_shot_results}
\centering
\footnotesize
\setlength{\tabcolsep}{2pt}

\begin{tabular}{l|c|c|c|c|c|c|c|c|c}
    \toprule
    & & \multicolumn{2}{c|}{Driving performance}
    & Efficiency
    & \multicolumn{4}{c|}{Geodesic accuracy}
    & Reliability \\
    
    Method
    & Split
    & EPDMS$\uparrow$
    & EPDMS$^\dagger\uparrow$
    & Time $\downarrow$
    & FDE $\downarrow$
    & $\Delta x$ $\downarrow$
    & $\Delta y$ $\downarrow$
    & $\Delta \theta$ $\downarrow$
    & Hit rate $\uparrow$ \\
    
    \midrule
    \multicolumn{10}{c}{100 subsampled test scenes} \\
    \midrule
    
    LeWM
    & \multirow{5}{*}{\texttt{navtrain}}
    & 48.3
    & 73.9
    & \textbf{0.7}
    & 12.4
    & 11.3
    & 2.7
    & 13.6
    & 6/18\\

    DINO-WM
    & 
    & 68.3
    & 91.4
    & 91.8
    & 3.9
    & 3.4
    & 1.1
    & 5.9
    & 40/73\\

    JEPA-WM
    & 
    & 74.2
    & 90.9
    & 101.0
    & 4.0
    & 3.4
    & 1.2
    & 5.2
    & 45/75\\

    AD-E2E-JEPA
    & 
    & \textbf{76.6}
    & \textbf{92.4}
    & 0.8
    & 4.2
    & 3.9
    & \textbf{1.0}
    & 4.7
    & 27/59\\

    + rollout
    &
    & 70.4
    & 92.2
    & 0.8
    & \textbf{3.5}
    & \textbf{3.2}
    & \textbf{1.0}
    & \textbf{4.1}
    & 34/67\\

    \midrule

    AD-E2E-JEPA
    & \multirow{2}{*}{\texttt{trainval}}
    & 72.1
    & 89.8
    & 0.8
    & 4.7
    & 4.4
    & 0.9
    & \textbf{3.4}
    & 33/55\\

    + rollout
    & 
    & 72.3
    & 91.9
    & 0.8
    & \textbf{3.2}
    & \textbf{3.0}
    & \textbf{0.7}
    & 3.6
    & \textbf{47}/\textbf{78}\\

    \midrule
    \midrule
    \multicolumn{10}{c}{Full 12,146 test scenes} \\
    \midrule
    
    LeWM
    & \multirow{3}{*}{\texttt{navtrain}}
    & 39.8
    & 66.7
    & \textbf{0.7}
    & 14.6
    & 13
    & 3.5
    & 17.2
    & 5.7/11.3\\

    AD-E2E-JEPA
    & 
    & 63.5
    & 80.1
    & 0.8
    & 6.3
    & 5.9
    & 1.2
    & 6.3
    & 32.9/65.3\\

    + rollout
    &
    & 64.9
    & 83.1
    & 0.8
    & 4.5
    & 4.0
    & 1.2
    & 4.1 
    & 42.6/71.0\\

    \midrule

    AD-E2E-JEPA
    & \multirow{7}{*}{\texttt{trainval}}
    & 63.2
    & 80.1
    & 0.8
    & 6.2
    & 5.8
    & 1.1
    & 4.5
    & 31.0/64.8\\

    + rollout
    &
    & 67.3
    & 84.1
    & 0.8
    & 4.0
    & 3.6
    & 1.1
    & 3.5
    & \textbf{53.8}/\textbf{82.7}\\
    
    + rollout, 512 traj.
    & 
    & 69.2
    & 85.0
    & 1.4
    & 3.6
    & 3.3
    & 0.9
    & 2.9
    & 45.2/73.9\\

    + rollout, 1024 traj.
    &
    & 70.5
    & 85.5
    & 2.5
    & 3.2
    & 2.9
    & \textbf{0.8}
    & 2.5
    & 36.3/64.3\\

    + rollout, 2048 traj.
    &
    & 71.5
    & 86.0
    & 4.7
    & 3.0
    & 2.7
    & \textbf{0.8}
    & 2.2
    & 27.4/53.2\\

    + rollout, 4096 traj.
    & 
    & 72.1
    & 86.3
    & 9.3
    & 2.9
    & 2.6
    & \textbf{0.8}
    & 2.1
    & 20.7/43.3\\

    + rollout, 8192 traj.
    & 
    & \textbf{72.9}
    & \textbf{86.5}
    & 18.2
    & \textbf{2.8}
    & \textbf{2.5}
    & \textbf{0.8}
    & \textbf{2.0}
    & 15.3/33.8\\
    
    \bottomrule
\end{tabular}

\end{table*}

As shown in Table~\ref{tab:zero_shot_results}, LeWM is computationally efficient but achieves substantially lower driving performance, geodesic accuracy, and hit rates than the other methods. DINO-WM and JEPA-WM are considerably more reliable, achieving higher hit rates and EPDMS scores of 68.3 and 74.2, respectively, on the 100-scene subset. However, their per-scene planning times are 91.8 and 101.0 seconds, respectively, making full-test-set evaluation prohibitively expensive. In contrast, AD-E2E-JEPA requires only 0.8 seconds per scene. When trained on \texttt{navtrain}, AD-E2E-JEPA achieves the highest driving performance on the 100-scene subset, although its geodesic accuracy and hit rate are lower than those of DINO-WM and JEPA-WM.

Adding the rollout loss and training on the larger \texttt{trainval} split substantially improves geodesic accuracy and reliability. On the 100-scene subset, this configuration achieves an FDE of 3.2 m and a top-1/top-5 hit rate of 47\%/78\%, comparable to or better than DINO-WM and JEPA-WM on these metrics. Its EPDMS is lower on this subset, which may partly reflect variance from evaluating only 100 scenes. This gap is reduced when evaluating on the full test set. Another possible explanation is that the world model is trained using a latent-distance objective, which is more directly aligned with geodesic distance and hit rate than with the NAVSIM-v2 driving metrics.

On the full 12,146-scene test set, AD-E2E-JEPA trained on \texttt{trainval} with rollout loss achieves an EPDMS of 67.3 and an EPDMS$^\dagger$ of 84.1. It also achieves an FDE of 4.0 m and a top-1/top-5 hit rate of 54\%/83\%, providing the strongest overall results on the full test set and significantly outperforms LeWM in the same setting. When we further scales the subsampled candidates to 512, 1024, 2048, 4096, 8192, it further boosts the performance and 8192 variant reaches an EPDMS of 72.9 and an EPDMS$^\dagger$ of 86.5 with FDE of 2.8 m. Hit rate decreases accordingly as the number of candidates increases.

\subsection{Transfer Learning Performance}

Recent high-performing E2E autonomous driving methods can require substantial training compute. For example, WA-JEPA~\citep{wang2026wa}, which reports state-of-the-art NAVSIM-v2 performance at submission time, uses 64 A800 GPUs for Stage~1 pretraining and 32 A800 GPUs for Stage~2 training. Rather than pursuing gains through large-scale training, we study a complementary question: whether the lightweight projector learned by self-supervised AD-E2E-JEPA world modeling transfers useful predictive structure to downstream imitation learning. We isolate the effect of projector pretraining without additional self-supervised pretraining of the encoder backbone. While prior JEPA-based E2E driving methods mainly study pretrained encoder representations or jointly pretrained world-action models, we are not aware of prior work that specifically isolates the downstream transferability of a pretrained projector. We include Transfuser~\citep{chitta2022transfuser}, Latent-WAM~\citep{wang2026latent}, Drive-JEPA~\citep{wang2026drive}, and WA-JEPA~\citep{wang2026wa} as reference methods rather than direct baselines.

\begin{table*}[ht!]
    \caption{
    NAVSIMv2 \texttt{navtest-stage-1} benchmark. V: single-view (SV) or multi-view (MV); Type: perception-free (PF) or perception-based (PB); Fr.: number of input frames. PB methods use perception annotations or trajectory-score labels. $^*$ denotes results obtained before the human-filter bug fix in commit \texttt{359c7f7} in NAVSIM's codebase.}
    \label{tab:benchmark_navsim_v2_stage1}
    \centering
    \footnotesize
    \setlength{\tabcolsep}{0.9pt}
    \renewcommand{\arraystretch}{1.0}

    \begin{tabular}{@{}lccc|ccccccccccc@{}}
    \toprule
    & & & &
    \multicolumn{11}{c}{NAVSIMv2 stage 1 driving metrics} \\
    Method & V & Type & Fr.
    & NC$\uparrow$
    & DAC$\uparrow$
    & DDC$\uparrow$
    & TLC$\uparrow$
    & \textbf{EP}$\uparrow$
    & TTC$\uparrow$
    & LK$\uparrow$
    & HC$\uparrow$
    & EC$\uparrow$
    & EPDMS$^*\uparrow$
    & EPDMS$\uparrow$ \\
    \midrule

    Transfuser
    & MV & PF & 1
    & 96.9 & 89.9 & 97.8 & 99.7 & 87.1
    & 95.4 & 92.7 & \textbf{98.3} & 87.2 & 76.7 & -- \\

    Latent-WAM
    & MV & PF & 4
    & 98.1 & 97.3 & 99.6 & 99.8 & 87.7
    & 97.3 & 97.6 & 98.1 & 72.4 & -- & 89.3 \\

    WA-JEPA
    & MV & PF & 4
    & \textbf{99.4} & 98.2 & \textbf{99.7} & \textbf{99.9} & 87.8
    & \textbf{98.9} & \textbf{98.3} & \textbf{98.3} & \textbf{88.1}
    & \textbf{88.0} & \textbf{91.7} \\

    Drive-JEPA
    & SV & PB & 2
    & 98.4 & \textbf{98.6} & 99.1 & 99.8 & \textbf{88.4}
    & 97.8 & 97.6 & 97.9 & 84.8 & 87.8 & -- \\

    \midrule
    \multicolumn{15}{@{}l}{DINOv3} \\[-1pt]

    + rand.\ proj.
    & SV & PF & 4
    & 96.8 & 89.8 & 98.3 & 99.7 & 87.1
    & 95.7 & 94.8 & 98.3 & 84.1 & -- & 80.2 \\

    + AD-E2E-JEPA proj.
    & SV & PF & 4
    & \textbf{97.7} & \textbf{93.7} & \textbf{99.2} & \textbf{99.8} & \textbf{87.3}
    & \textbf{96.8} & \textbf{97.0} & \textbf{98.4} & \textbf{88.7} & --
    & \textbf{85.4} \\

    \bottomrule
    \end{tabular}
\end{table*}

\section{Conclusion}

We propose AD-E2E-JEPA, a joint-embedding predictive architecture for end-to-end autonomous driving. To isolate the effect of world modeling from driving policy learning, we perform zero-shot goal-conditioned planning and quantitatively evaluate the JEPA-based world model in terms of driving performance, planning efficiency, geodesic accuracy, and  reliability. We further introduce a learnable projector with SIGReg to compress latent representations for planning while preserving their information content. Without any policy training, AD-E2E-JEPA improves planning efficiency by $100\times$ compared with DINO-WM/JEPA-WM under the same setting of 256 candidate trajectories. The best AD-E2E-JEPA variant, using 8192 candidate trajectories, achieves an EPDMS of 72.9, reaches goal with a displacement error of 2.8 meters and a mean heading error of 2.0 degrees. It also demonstrates the projector pretrained during world modeling benefits imitation learning.

\bibliography{iclr2027_conference}

@article{oquab2023dinov2,
  title={Dinov2: Learning robust visual features without supervision},
  author={Oquab, Maxime and Darcet, Timoth{\'e}e and Moutakanni, Th{\'e}o and Vo, Huy and Szafraniec, Marc and Khalidov, Vasil and Fernandez, Pierre and Haziza, Daniel and Massa, Francisco and El-Nouby, Alaaeldin and others},
  journal={arXiv preprint arXiv:2304.07193},
  year={2023}
}

@article{cao2025pseudo,
  title={Pseudo-simulation for autonomous driving},
  author={Cao, Wei and Hallgarten, Marcel and Li, Tianyu and Dauner, Daniel and Gu, Xunjiang and Wang, Caojun and Miron, Yakov and Aiello, Marco and Li, Hongyang and Gilitschenski, Igor and others},
  journal={arXiv preprint arXiv:2506.04218},
  year={2025}
}

@article{dauner2024navsim,
  title={Navsim: Data-driven non-reactive autonomous vehicle simulation and benchmarking},
  author={Dauner, Daniel and Hallgarten, Marcel and Li, Tianyu and Weng, Xinshuo and Huang, Zhiyu and Yang, Zetong and Li, Hongyang and Gilitschenski, Igor and Ivanovic, Boris and Pavone, Marco and others},
  journal={Advances in Neural Information Processing Systems},
  volume={37},
  pages={28706--28719},
  year={2024}
}

@article{su2024roformer,
  title={Roformer: Enhanced transformer with rotary position embedding},
  author={Su, Jianlin and Ahmed, Murtadha and Lu, Yu and Pan, Shengfeng and Bo, Wen and Liu, Yunfeng},
  journal={Neurocomputing},
  volume={568},
  pages={127063},
  year={2024},
  publisher={Elsevier}
}

@article{zhou2024dino,
  title={Dino-wm: World models on pre-trained visual features enable zero-shot planning},
  author={Zhou, Gaoyue and Pan, Hengkai and LeCun, Yann and Pinto, Lerrel},
  journal={arXiv preprint arXiv:2411.04983},
  year={2024}
}

@article{li2024hydra,
  title={Hydra-mdp: End-to-end multimodal planning with multi-target hydra-distillation},
  author={Li, Zhenxin and Li, Kailin and Wang, Shihao and Lan, Shiyi and Yu, Zhiding and Ji, Yishen and Li, Zhiqi and Zhu, Ziyue and Kautz, Jan and Wu, Zuxuan and others},
  journal={arXiv preprint arXiv:2406.06978},
  year={2024}
}

@article{li2025hydra,
  title={Hydra-mdp++: Advancing end-to-end driving via expert-guided hydra-distillation},
  author={Li, Kailin and Li, Zhenxin and Lan, Shiyi and Xie, Yuan and Zhang, Zhizhong and Liu, Jiayi and Wu, Zuxuan and Yu, Zhiding and Alvarez, Jose M},
  journal={arXiv preprint arXiv:2503.12820},
  year={2025}
}

@article{wang2026latent,
  title={Latent-WAM: Latent World Action Modeling for End-to-End Autonomous Driving},
  author={Wang, Linbo and Zheng, Yupeng and Chen, Qiang and Li, Shiwei and Zhang, Yichen and Xing, Zebin and Zhang, Qichao and Li, Xiang and Qian, Deheng and Yang, Pengxuan and others},
  journal={arXiv preprint arXiv:2603.24581},
  year={2026}
}

@article{terver2025drives,
  title={What Drives Success in Physical Planning with Joint-Embedding Predictive World Models?},
  author={Terver, Basile and Yang, Tsung-Yen and Ponce, Jean and Bardes, Adrien and LeCun, Yann},
  journal={arXiv preprint arXiv:2512.24497},
  year={2025}
}

@article{simeoni2025dinov3,
  title={Dinov3},
  author={Sim{\'e}oni, Oriane and Vo, Huy V and Seitzer, Maximilian and Baldassarre, Federico and Oquab, Maxime and Jose, Cijo and Khalidov, Vasil and Szafraniec, Marc and Yi, Seungeun and Ramamonjisoa, Micha{\"e}l and others},
  journal={arXiv preprint arXiv:2508.10104},
  year={2025}
}

@article{assran2025v,
  title={V-jepa 2: Self-supervised video models enable understanding, prediction and planning},
  author={Assran, Mido and Bardes, Adrien and Fan, David and Garrido, Quentin and Howes, Russell and Muckley, Matthew and Rizvi, Ammar and Roberts, Claire and Sinha, Koustuv and Zholus, Artem and others},
  journal={arXiv preprint arXiv:2506.09985},
  year={2025}
}

@inproceedings{perez2018film,
  title={Film: Visual reasoning with a general conditioning layer},
  author={Perez, Ethan and Strub, Florian and De Vries, Harm and Dumoulin, Vincent and Courville, Aaron},
  booktitle={Proceedings of the AAAI conference on artificial intelligence},
  year={2018}
}

@article{chen2024end,
  title={End-to-end autonomous driving: Challenges and frontiers},
  author={Chen, Li and Wu, Penghao and Chitta, Kashyap and Jaeger, Bernhard and Geiger, Andreas and Li, Hongyang},
  journal={IEEE Transactions on Pattern Analysis and Machine Intelligence},
  year={2024},
  publisher={IEEE}
}

@inproceedings{hu2023planning,
  title={Planning-oriented autonomous driving},
  author={Hu, Yihan and Yang, Jiazhi and Chen, Li and Li, Keyu and Sima, Chonghao and Zhu, Xizhou and Chai, Siqi and Du, Senyao and Lin, Tianwei and Wang, Wenhai and others},
  booktitle={Proceedings of the IEEE/CVF conference on computer vision and pattern recognition},
  pages={17853--17862},
  year={2023}
}

@article{lecun2022path,
  title={A path towards autonomous machine intelligence version 0.9. 2, 2022-06-27},
  author={LeCun, Yann and others},
  journal={Open Review},
  volume={62},
  number={1},
  pages={1--62},
  year={2022}
}

@article{li2024enhancing,
  title={Enhancing end-to-end autonomous driving with latent world model},
  author={Li, Yingyan and Fan, Lue and He, Jiawei and Wang, Yuqi and Chen, Yuntao and Zhang, Zhaoxiang and Tan, Tieniu},
  journal={arXiv preprint arXiv:2406.08481},
  year={2024}
}

@article{li2024navigation,
  title={Navigation-guided sparse scene representation for end-to-end autonomous driving},
  author={Li, Peidong and Cui, Dixiao},
  journal={arXiv preprint arXiv:2409.18341},
  year={2024}
}

@article{balestriero2025lejepa,
  title={Lejepa: Provable and scalable self-supervised learning without the heuristics},
  author={Balestriero, Randall and LeCun, Yann},
  journal={arXiv preprint arXiv:2511.08544},
  year={2025}
}

@article{yang2026auto,
  title={Auto-JEPA: A Latent World Model of Continuous Intent for End-to-End Autonomous Driving},
  author={Yang, Jiwei and Chen, Zhengxian and Huang, Chaosheng and Li, Jun},
  journal={arXiv preprint arXiv:2607.29031},
  year={2026}
}

@article{zhong2026wam,
  title={DA-WAM: Decision-Aligned Future Latents for Driving World Models},
  author={Zhong, Ruiguo and Ma, Benshan and Chen, Xiaolong and Zhang, Lang and Feng, Mingyue and Wang, Yaonong and Liu, Pei and Ma, Jun},
  journal={arXiv preprint arXiv:2608.19085},
  year={2026}
}

@article{wang2026wa,
  title={WA-JEPA: Rethinking the Video JEPA Paradigm for World-Action Modeling in Autonomous Driving},
  author={Wang, Xinlin and Xiang, Yujiao and Zhou, Yuheng and Wang, Jingqi and Huang, Minqing and Huang, Jiajie and Wei, Dongxu and Zhou, Tingguang and Wang, Xiyang and Chen, Gong and others},
  journal={arXiv preprint arXiv:2608.20974},
  year={2026}
}

@article{wang2026drive,
  title={Drive-jepa: Video jepa meets multimodal trajectory distillation for end-to-end driving},
  author={Wang, Linhan and Yang, Zichong and Bai, Chen and Zhang, Guoxiang and Liu, Xiaotong and Zheng, Xiaoyin and Long, Xiao-Xiao and Lu, Chang-Tien and Lu, Cheng},
  journal={arXiv preprint arXiv:2601.22032},
  year={2026}
}

@inproceedings{zhu2026self,
  title={Self-supervised representation learning with joint embedding predictive architecture for automotive lidar object detection},
  author={Zhu, Haoran and Dong, Zhenyuan and Topollai, Kristi and Sha, Beiyao and Choromanska, Anna Ewa},
  booktitle={Proceedings of the AAAI Conference on Artificial Intelligence},
  year={2026}
}

@article{naeinian2026zero,
  title={Zero-Shot Cross-City Generalization in End-to-End Autonomous Driving: Self-Supervised versus Supervised Representations},
  author={Naeinian, Fatemeh and Hamza, Ali and Zhu, Haoran and Choromanska, Anna},
  journal={arXiv preprint arXiv:2603.11417},
  year={2026}
}

@inproceedings{peebles2023scalable,
  title={Scalable diffusion models with transformers},
  author={Peebles, William and Xie, Saining},
  booktitle={2023 IEEE/CVF International Conference on Computer Vision (ICCV)},
  pages={4172--4182},
  year={2023},
  organization={IEEE}
}

@inproceedings{zhang2025epona,
  title={Epona: Autoregressive diffusion world model for autonomous driving},
  author={Zhang, Kaiwen and Tang, Zhenyu and Hu, Xiaotao and Pan, Xingang and Guo, Xiaoyang and Liu, Yuan and Huang, Jingwei and Yuan, Li and Zhang, Qian and Long, Xiao-Xiao and others},
  booktitle={2025 IEEE/CVF International Conference on Computer Vision (ICCV)},
  pages={27220--27230},
  year={2025},
  organization={IEEE}
}

@inproceedings{
wang2026temporal,
title={Temporal Straightening for Latent Planning},
author={Ying Wang and Oumayma Bounou and Gaoyue Zhou and Randall Balestriero and Tim G. J. Rudner and Yann LeCun and Mengye Ren},
booktitle={Forty-third International Conference on Machine Learning},
year={2026},
url={https://openreview.net/forum?id=Ik1mKtUYlZ}
}

@article{maes2026leworldmodel,
  title={Leworldmodel: Stable end-to-end joint-embedding predictive architecture from pixels},
  author={Maes, Lucas and Lidec, Quentin Le and Scieur, Damien and LeCun, Yann and Balestriero, Randall},
  journal={arXiv preprint arXiv:2603.19312},
  year={2026}
}

@article{kuhn2026levjepa,
  title={LeVJEPA: Efficient \& Scalable Video Pretraining without the Heuristics},
  author={Kuhn, Lukas and Maes, Lucas and Serra, Giuseppe and Lidec, Quentin Le and LeCun, Yann and Balestriero, Randall and Buettner, Florian},
  journal={arXiv preprint arXiv:2608.27395},
  year={2026}
}

@article{epps1983test,
  title={A test for normality based on the empirical characteristic function},
  author={Epps, Thomas W and Pulley, Lawrence B},
  journal={Biometrika},
  volume={70},
  number={3},
  pages={723--726},
  year={1983},
  publisher={Oxford University Press}
}

@article{cramer1936some,
  title={Some theorems on distribution functions},
  author={Cram{\'e}r, Harald and Wold, Herman},
  journal={Journal of the London Mathematical Society},
  volume={1},
  number={4},
  pages={290--294},
  year={1936},
  publisher={Wiley Online Library}
}

@article{sobal2026learning,
  title={Learning from reward-free offline data: A case for planning with latent dynamics models},
  author={Sobal, Uladzislau and Zhang, Wancong and Cho, Kyunghyun and Balestriero, Randall and Rudner, Tim GJ and LeCun, Yann},
  journal={Advances in Neural Information Processing Systems},
  volume={38},
  pages={43905--43941},
  year={2026}
}

@article{chen2024vadv2,
  title={Vadv2: End-to-end vectorized autonomous driving via probabilistic planning},
  author={Chen, Shaoyu and Jiang, Bo and Gao, Hao and Liao, Bencheng and Xu, Qing and Zhang, Qian and Huang, Chang and Liu, Wenyu and Wang, Xinggang},
  journal={arXiv preprint arXiv:2402.13243},
  year={2024}
}

@inproceedings{kirby2026driving,
  title={Driving on registers},
  author={Kirby, Ellington and Boulch, Alexandre and Xu, Yihong and Yin, Yuan and Puy, Gilles and Zablocki, {\'E}loi and Bursuc, Andrei and Gidaris, Spyros and Marlet, Renaud and Bartoccioni, Florent and others},
  booktitle={Proceedings of the IEEE/CVF Conference on Computer Vision and Pattern Recognition},
  pages={32058--32069},
  year={2026}
}

@inproceedings{finn2017deep,
  title={Deep visual foresight for planning robot motion},
  author={Finn, Chelsea and Levine, Sergey},
  booktitle={2017 IEEE international conference on robotics and automation (ICRA)},
  pages={2786--2793},
  year={2017},
  organization={IEEE}
}

@article{sutton1991dyna,
  title={Dyna, an integrated architecture for learning, planning, and reacting},
  author={Sutton, Richard S},
  journal={ACM Sigart Bulletin},
  volume={2},
  number={4},
  pages={160--163},
  year={1991},
  publisher={ACM New York, NY, USA}
}

@book{bryson1975applied,
  title={Applied optimal control: optimization, estimation and control},
  author={Bryson, Arthur Earl},
  year={1975},
  publisher={CRC press}
}

@article{kang2024far,
  title={How far is video generation from world model: A physical law perspective},
  author={Kang, Bingyi and Yue, Yang and Lu, Rui and Lin, Zhijie and Zhao, Yang and Wang, Kaixin and Huang, Gao and Feng, Jiashi},
  journal={arXiv preprint arXiv:2411.02385},
  year={2024}
}

@article{ha2018world,
  title={World models},
  author={Ha, David and Schmidhuber, J{\"u}rgen},
  journal={arXiv preprint arXiv:1803.10122},
  volume={2},
  number={3},
  pages={440},
  year={2018}
}

@inproceedings{mur2026v,
  title={V-jepa 2.1: Unlocking dense features in video self-supervised learning},
  author={Mur-Labadia, Lorenzo and Muckley, Matthew and Bar, Amir and Assran, Mido and Sinha, Koustuv and Rabbat, Mike and LeCun, Yann and Ballas, Nicolas},
  booktitle={European Conference on Computer Vision},
  pages={671--689},
  year={2026},
  organization={Springer}
}

@article{wang2026music,
  title={Music-JEPA: Learning a World Model of Sound from Action},
  author={Wang, Ziyu and Fang, Kun and LeCun, Yann},
  journal={arXiv preprint arXiv:2607.22000},
  year={2026}
}

@article{fei2023jepa,
  title={A-jepa: Joint-embedding predictive architecture can listen},
  author={Fei, Zhengcong and Fan, Mingyuan and Huang, Junshi},
  journal={arXiv preprint arXiv:2311.15830},
  year={2023}
}

@article{bardes2021vicreg,
  title={Vicreg: Variance-invariance-covariance regularization for self-supervised learning},
  author={Bardes, Adrien and Ponce, Jean and LeCun, Yann},
  journal={arXiv preprint arXiv:2105.04906},
  year={2021}
}

@article{wu2026visreg,
  title={VISReg: Variance-Invariance-Sketching Regularization for JEPA training},
  author={Wu, Haiyu and Balestriero, Randall and Levine, Morgan},
  journal={arXiv preprint arXiv:2606.02572},
  year={2026}
}

@article{kuang2026lpwm,
  title={LpWM: A Case for Sparse Representations in World Models},
  author={Kuang, Yilun and Dagade, Yash and Lidec, Quentin Le and Maes, Lucas and Balestriero, Randall and LeCun, Yann},
  journal={arXiv preprint arXiv:2608.22764},
  year={2026}
}

@article{zhang2026hierarchical,
  title={Hierarchical planning with latent world models},
  author={Zhang, Wancong and Terver, Basile and Zholus, Artem and Chitnis, Soham and Sutaria, Harsh and Assran, Mido and Balestriero, Randall and Bar, Amir and Bardes, Adrien and LeCun, Yann and others},
  journal={arXiv preprint arXiv:2604.03208},
  year={2026}
}

@article{kuang2026rectified,
  title={Rectified LpJEPA: Joint-Embedding Predictive Architectures with Sparse and Maximum-Entropy Representations},
  author={Kuang, Yilun and Dagade, Yash and Rudner, Tim GJ and Balestriero, Randall and LeCun, Yann},
  journal={arXiv preprint arXiv:2602.01456},
  year={2026}
}

@article{garrido2025intuitive,
  title={Intuitive physics understanding emerges from self-supervised pretraining on natural videos},
  author={Garrido, Quentin and Ballas, Nicolas and Assran, Mahmoud and Bardes, Adrien and Najman, Laurent and Rabbat, Michael and Dupoux, Emmanuel and LeCun, Yann},
  journal={arXiv preprint arXiv:2502.11831},
  year={2025}
}

@inproceedings{assran2023self,
  title={Self-supervised learning from images with a joint-embedding predictive architecture},
  author={Assran, Mahmoud and Duval, Quentin and Misra, Ishan and Bojanowski, Piotr and Vincent, Pascal and Rabbat, Michael and LeCun, Yann and Ballas, Nicolas},
  booktitle={2023 IEEE/CVF Conference on Computer Vision and Pattern Recognition (CVPR)},
  pages={15619--15629},
  year={2023},
  organization={IEEE}
}

@article{bardes2024revisiting,
  title={Revisiting feature prediction for learning visual representations from video},
  author={Bardes, Adrien and Garrido, Quentin and Ponce, Jean and Chen, Xinlei and Rabbat, Michael and LeCun, Yann and Assran, Mahmoud and Ballas, Nicolas},
  journal={arXiv preprint arXiv:2404.08471},
  year={2024}
}

@article{chitta2022transfuser,
  title={Transfuser: Imitation with transformer-based sensor fusion for autonomous driving},
  author={Chitta, Kashyap and Prakash, Aditya and Jaeger, Bernhard and Yu, Zehao and Renz, Katrin and Geiger, Andreas},
  journal={IEEE transactions on pattern analysis and machine intelligence},
  volume={45},
  number={11},
  pages={12878--12895},
  year={2022},
  publisher={IEEE}
}

@article{wang2026adajepa,
  title={AdaJEPA: An Adaptive Latent World Model},
  author={Wang, Ying and Bounou, Oumayma and LeCun, Yann and Ren, Mengye},
  journal={arXiv preprint arXiv:2606.32026},
  year={2026}
}

@inproceedings{jiang2023vad,
  title={Vad: Vectorized scene representation for efficient autonomous driving},
  author={Jiang, Bo and Chen, Shaoyu and Xu, Qing and Liao, Bencheng and Chen, Jiajie and Zhou, Helong and Zhang, Qian and Liu, Wenyu and Huang, Chang and Wang, Xinggang},
  booktitle={2023 IEEE/CVF International Conference on Computer Vision (ICCV)},
  pages={8306--8316},
  year={2023},
  organization={IEEE}
}

@article{hu2023gaia,
  title={Gaia-1: A generative world model for autonomous driving},
  author={Hu, Anthony and Russell, Lloyd and Yeo, Hudson and Murez, Zak and Fedoseev, George and Kendall, Alex and Shotton, Jamie and Corrado, Gianluca},
  journal={arXiv preprint arXiv:2309.17080},
  year={2023}
}

@article{bojarski2016end,
  title={End to end learning for self-driving cars},
  author={Bojarski, Mariusz and Del Testa, Davide and Dworakowski, Daniel and Firner, Bernhard and Flepp, Beat and Goyal, Prasoon and Jackel, Lawrence D and Monfort, Mathew and Muller, Urs and Zhang, Jiakai and others},
  journal={arXiv preprint arXiv:1604.07316},
  year={2016}
}

@article{pomerleau1988alvinn,
  title={Alvinn: An autonomous land vehicle in a neural network},
  author={Pomerleau, Dean A},
  journal={Advances in neural information processing systems},
  volume={1},
  year={1988}
}

@inproceedings{liao2025diffusiondrive,
  title={Diffusiondrive: Truncated diffusion model for end-to-end autonomous driving},
  author={Liao, Bencheng and Chen, Shaoyu and Yin, Haoran and Jiang, Bo and Wang, Cheng and Yan, Sixu and Zhang, Xinbang and Li, Xiangyu and Zhang, Ying and Zhang, Qian and others},
  booktitle={2025 IEEE/CVF Conference on Computer Vision and Pattern Recognition (CVPR)},
  pages={12037--12047},
  year={2025},
  organization={IEEE}
}

@inproceedings{chen2021exploring,
  title={Exploring simple siamese representation learning},
  author={Chen, Xinlei and He, Kaiming},
  booktitle={2021 IEEE/CVF conference on computer vision and pattern recognition (CVPR)},
  pages={15745--15753},
  year={2021},
  organization={IEEE}
}

@article{grill2020bootstrap,
  title={Bootstrap your own latent-a new approach to self-supervised learning},
  author={Grill, Jean-Bastien and Strub, Florian and Altch{\'e}, Florent and Tallec, Corentin and Richemond, Pierre and Buchatskaya, Elena and Doersch, Carl and Avila Pires, Bernardo and Guo, Zhaohan and Gheshlaghi Azar, Mohammad and others},
  journal={Advances in neural information processing systems},
  volume={33},
  pages={21271--21284},
  year={2020}
}

@book{jaeger2002tutorial,
  title={Tutorial on training recurrent neural networks, covering BPPT, RTRL, EKF and the echo state network approach},
  author={Jaeger, Herbert},
  year={2002},
  publisher={GMD-Forschungszentrum Informationstechnik Bonn}
}

@article{zhu2026adlistjepa,
  title={Self-Supervised JEPA-based World Models for LiDAR Occupancy Completion and Forecasting},
  author={Zhu, Haoran and Choromanska, Anna},
  journal={arXiv preprint arXiv:2602.12540},
  year={2026}
}
\bibliographystyle{iclr2027_conference}

\appendix
\section{Appendix}
\subsection{SIGReg Configuration Details}
\label{appendix:sigreg}

$T(\cdot)$ denotes the univariate Epps--Pulley test~\citep{epps1983test}. We use the default SIGReg settings~\citep{balestriero2025lejepa} used in LeWorld~\citep{maes2026leworldmodel,kuhn2026levjepa}, where the integral is approximated with 17 knots over $[0, 3]$ and $M=1024$ directions.

\subsection{Rollout Setting}
\label{appendix:loss_with_rollout}

For the AD-E2E-JEPA variant with rollout training in
Eq.~(\ref{eq:overall_loss_with_rollout}), we adapt the rollout
implementation of~\citet{terver2025drives} to the full future horizon of
$F$ frames. Each prediction window contains $K+2$ consecutive frames:
the first $K+1$ frames are used as input, and the predictor produces the
corresponding one-step-shifted predictions. In the NAVSIM setting,
$F=8$ and $K+1=4$.

\subsubsection{Teacher-Forcing Loss}

The first component is the teacher-forcing prediction loss
$\overline{\mathcal{L}}_{\mathrm{TF}}$. For each time window, we use the
ground-truth context embeddings as input, without autoregressive
rollout. In our current implementation: for $k=0$, we supervise the entire prediction window; For each
subsequent teacher-forced prediction ($k\geq 1$), we supervise only the
final time step of the prediction window. Specifically,
\begin{equation}
    \hat{z}^{\mathrm{TF}}_{t+k-K+1:t+k+1}
    =
    \operatorname{Pred}^{\prime}\!\left(
        z_{t+k-K:t+k},
        E_a(\mathbf{a}_{t+k-K:t+k})
    \right),
    \qquad
    k=0,\ldots,F-1.
\end{equation}

For the first prediction window,
\begin{equation}
    \mathcal{L}^{\mathrm{proj}}_{\mathrm{pred}}[0]
    =
    \operatorname{MSE}\!\left(
        \hat{z}^{\mathrm{TF}}_{t-K+1:t+1},
        \operatorname{sg}\!\left(z_{t-K+1:t+1}\right)
    \right).
\end{equation}

For the subsequent prediction windows,
\begin{equation}
    \mathcal{L}^{\mathrm{proj}}_{\mathrm{pred}}[k]
    =
    \operatorname{MSE}\!\left(
        \hat{z}^{\mathrm{TF}}_{t+k+1},
        \operatorname{sg}\!\left(z_{t+k+1}\right)
    \right),
    \qquad
    k=1,\ldots,F-1.
\end{equation}

The teacher-forcing loss is then
\begin{equation}
    \overline{\mathcal{L}}_{\mathrm{TF}}
    =
    \frac{
        (K+1)\,\mathcal{L}^{\mathrm{proj}}_{\mathrm{pred}}[0]
        +
        \sum_{k=1}^{F-1}
        \mathcal{L}^{\mathrm{proj}}_{\mathrm{pred}}[k]
    }{
        K+F
    }.
\end{equation}

\subsubsection{Rollout Consistency Loss}

The second component consists of the rollout consistency losses
$\overline{\mathcal{L}}_{k}$ for future horizons $k=2,\ldots,F$.
We perform a single autoregressive rollout of $F$ steps starting
from the initial context window $z_{t-K:t}$. Throughout the rollout,
we maintain a context window of $K+1$ embeddings. After each prediction,
we remove the oldest context embedding and append only the final
predicted embedding.

Specifically, we initialize the autoregressive context as
\begin{equation}
    \tilde{z}^{\mathrm{AR},(0)}_{t-K:t}
    =
    z_{t-K:t}.
\end{equation}

We reuse the initial teacher-forcing prediction as the first
autoregressive prediction, which uses the initial context without
stop-gradient:
\begin{equation}
    \hat{z}^{\mathrm{AR},(1)}_{t-K+1:t+1}
    =
    \operatorname{Pred}^{\prime}\!\left(
        \tilde{z}^{\mathrm{AR},(0)}_{t-K:t},
        E_a(\mathbf{a}_{t-K:t})
    \right).
\end{equation}

Following~\citet{terver2025drives}, we apply truncated backpropagation
through time (TBPTT)~\citep{jaeger2002tutorial} by stopping gradients
through the autoregressive context before each subsequent predictor
call. For rollout steps $j=2,\ldots,F$, the predictor produces
\begin{equation}
    \hat{z}^{\mathrm{AR},(j)}_{t-K+j:t+j}
    =
    \operatorname{Pred}^{\prime}\!\left(
        \operatorname{sg}\!\left(
            \tilde{z}^{\mathrm{AR},(j-1)}_{t-K+j-1:t+j-1}
        \right),
        E_a(\mathbf{a}_{t-K+j-1:t+j-1})
    \right).
\end{equation}

After each prediction, the autoregressive context is updated by
removing its earliest embedding and appending the final predicted
embedding:
\begin{equation}
    \tilde{z}^{\mathrm{AR},(j)}_{t-K+j:t+j}
    =
    \left[
        \tilde{z}^{\mathrm{AR},(j-1)}_{t-K+j:t+j-1},
        \hat{z}^{\mathrm{AR},(j)}_{t+j}
    \right],
    \qquad
    j=1,\ldots,F.
\end{equation}

The rollout consistency loss at horizon $k$ supervises only the
final prediction in the corresponding window against the ground-truth
embedding:
\begin{equation}
    \overline{\mathcal{L}}_{k}
    =
    \operatorname{MSE}\!\left(
        \hat{z}^{\mathrm{AR},(k)}_{t+k},
        \operatorname{sg}\!\left(z_{t+k}\right)
    \right),
    \qquad
    k=2,\ldots,F.
\end{equation}

The first-step prediction is supervised by the teacher-forcing loss
and is therefore excluded from the additional rollout loss terms.
Each horizon loss remains differentiable through its current predictor
call, while stop-gradient prevents backpropagation through earlier
rollout steps.

\subsubsection{SIGReg}

The third component is SIGReg, which is applied over all time steps
from $t-K$ through $t+F$, yielding
$\mathcal{L}^{t-K:t+F}_{\mathrm{SIGReg}}$.

Finally, the overall loss for the rollout setting is given by
Eq.~(\ref{eq:overall_loss_with_rollout}).

\subsection{Metrics}
\label{appendix:metrics}

\textbf{EPDMS.}
EPDMS is a pseudo-simulation-based metric for evaluating end-to-end autonomous driving planning, introduced in the NAVSIM benchmark~\citep{cao2025pseudo, dauner2024navsim}. It evaluates both safety and driving quality. Specifically, EPDMS consists of a multiplicative term capturing safety-related metrics, including no at-fault collision (NC), drivable area compliance (DAC), driving direction compliance (DDC), and traffic light compliance (TLC), and a weighted-average term capturing driving quality, including ego progress (EP), time-to-collision (TTC), lane keeping (LK), history comfort (HC), and extended comfort (EC). EC is included only for samples for which a valid neighboring scene is available.

We define an indicator $b_i$ denoting the availability of a valid neighboring scene:
\begin{equation}
b_i =
\begin{cases}
1, & \text{if a valid neighboring scene is available for computing EC},\\
0, & \text{otherwise}.
\end{cases}
\end{equation}
The EPDMS for sample $i$ is then computed as
\begin{equation}
\label{eq:epdms}
\mathrm{EPDMS}_i
=
\mathrm{NC}_i \cdot \mathrm{DAC}_i \cdot
\mathrm{DDC}_i \cdot \mathrm{TLC}_i
\cdot
\frac{
5\,\mathrm{EP}_i
+5\,\mathrm{TTC}_i
+2\,\mathrm{LK}_i
+2\,\mathrm{HC}_i
+2\,b_i\,\mathrm{EC}_i
}{
14+2\,b_i
}.
\end{equation}

\textbf{EPDMS$^\dagger$.}
Since zero-shot goal-conditioned planning with a world model is not directly optimized for the safety-related components of EPDMS, we additionally report EPDMS$^\dagger$, which excludes the multiplicative safety terms and evaluates driving quality using only the weighted-average component:
\begin{equation}
\label{eq:epdms_dagger}
\mathrm{EPDMS}^\dagger_i
=
\frac{
5\,\mathrm{EP}_i
+5\,\mathrm{TTC}_i
+2\,\mathrm{LK}_i
+2\,\mathrm{HC}_i
+2\,b_i\,\mathrm{EC}_i
}{
14+2\,b_i
}.
\end{equation}

\textbf{Planning time.}
We report the average per-scene planning time to assess the computational efficiency of end-to-end autonomous driving planning using a single NVIDIA A100 80 GB GPU, with all candidate trajectories evaluated in parallel on the GPU.

\textbf{Final-pose error (FDE, $\Delta x$, $\Delta y$, $\Delta \theta$).}
We evaluate the final-pose error between the ground-truth pose corresponding to the goal image and the final pose of the selected trajectory. FDE denotes the mean Euclidean distance between the ground-truth and selected final positions. We additionally report the mean absolute errors along the $x$- and $y$-axes, denoted by $\Delta x$ and $\Delta y$, respectively, as well as the mean absolute heading error $\Delta \theta$.

\textbf{Hit rate.}
Hit rate is a diagnostic metric introduced in the DrivoR~\citep{kirby2026driving} codebase, where it was originally used to evaluate whether the proposed scorer can identify high-scoring trajectories. Inspired by this idea, we adapt hit rate to evaluate the reliability of the world model. Given the subsampled candidate trajectory set $V_{\text{sampled}}$, we append the ground-truth trajectory to the candidate set and perform world-model rollouts for all trajectories. We then compute the planning cost for each trajectory according to Eq.~(\ref{eq:planning_cost}) and determine whether the ground-truth trajectory ranks among the top-$k$ trajectories with the lowest planning costs. Specifically, we report Top-1 and Top-5 hit rates, which measure the percentage of evaluation scenes in which the ground-truth trajectory is ranked among the top 1 and top 5 lowest-cost trajectories, respectively.

For each evaluation scene $n$, we define
\begin{equation}
h_n^{@k}
=
\mathbb{I}
\left[
\operatorname{rank}\!\left(C_n^{\mathrm{gt}}\right) \leq k
\right],
\end{equation}
where $C_n^{\mathrm{gt}}$ denotes the planning cost of the ground-truth trajectory, and the rank is computed over the planning costs of all trajectories in
$V_{\text{sampled}} \cup \{P_n^{\mathrm{gt}}\}$, with lower planning costs corresponding to higher ranks. The Top-$k$ hit rate is then
\begin{equation}
\label{eq:hit_rate}
\mathrm{HitRate}@k
=
\frac{1}{Q}
\sum_{n=1}^{Q}
h_n^{@k},
\qquad k \in \{1,5\},
\end{equation}
where $Q$ denotes the number of evaluation scenes.

\subsection{EPDMS Details for Zero-Shot Goal-Conditioned Planning}
\label{appendix:epdms_details}

Detailed EPDMS results are reported in Table~\ref{tab:zero_shot_epdms_details_100} for 100 subsampled test scenes and in Table~\ref{tab:zero_shot_epdms_details} for the full set of 12,146 test scenes.

\begin{table*}[ht!]
\caption{EPDMS details for 100 subsampled test scenes.}
\label{tab:zero_shot_epdms_details_100}
\centering
\footnotesize
\setlength{\tabcolsep}{1.0pt}

\begin{tabular}{l|c|c|c|c|c|c|c|c|c|c|c|c}
\toprule
Method
& Split
& NC$\uparrow$
& DAC$\uparrow$
& DDC$\uparrow$
& TLC$\uparrow$
& EP$\uparrow$
& TTC$\uparrow$
& LK$\uparrow$
& HC$\uparrow$
& EC$\uparrow$
& EPDMS$\uparrow$
& EPDMS$^\dagger$$\uparrow$ \\
\midrule

LeWM
& \multirow{5}{*}{\texttt{navtrain}}
& 78.5 & 77.0 & 84.0 & 97.0 & 68.6 & 76.0 & 86.0 & 70.0 & --
& 48.3
& 73.9 \\

DINO-WM
&
& \textbf{96.0} & 79.0 & 96.5 & 99.0 & 87.8 & 93.0 & 92.0 & \textbf{96.0} & --
& 68.3
& 91.4 \\

JEPA-WM
&
& 94.5 & \textbf{87.0} & 96.5 & 99.0 & 87.6 & 92.0 & 94.0 & 93.0 & --
& 74.2
& 90.9 \\

AD-E2E-JEPA
&
& \textbf{96.0} & \textbf{87.0} & 95.5 & 99.0 & \textbf{88.0} & \textbf{95.0} & 94.0 & 95.0 & --
& \textbf{76.6}
& \textbf{92.4} \\

\quad + rollout
&
& \textbf{96.0} & 77.0 & 95.5 & \textbf{100.0} & 87.9 & \textbf{95.0} & 93.0 & 95.0 & --
& 70.4
& 92.2 \\

\midrule

AD-E2E-JEPA
& \multirow{2}{*}{\texttt{trainval}}
& 94.0 & 82.0 & 96.5 & \textbf{100.0} & 87.2 & 92.0 & 91.0 & 90.0 & --
& 72.1
& 89.8 \\

\quad + rollout
&
& 95.5 & 82.0 & \textbf{98.0} & 99.0 & 87.3 & 93.0 & \textbf{97.0} & \textbf{96.0} & --
& 72.3
& 91.9 \\

\bottomrule
\end{tabular}
\end{table*}

\begin{table*}[ht!]
\caption{EPDMS details for the full set of 12,146 test scenes.}
\label{tab:zero_shot_epdms_details}
\centering
\footnotesize
\setlength{\tabcolsep}{0.9pt}

\begin{tabular}{l|c|c|c|c|c|c|c|c|c|c|c|c}
\toprule
Method
& Split
& NC$\uparrow$
& DAC$\uparrow$
& DDC$\uparrow$
& TLC$\uparrow$
& EP$\uparrow$
& TTC$\uparrow$
& LK$\uparrow$
& HC$\uparrow$
& EC$\uparrow$
& EPDMS$\uparrow$
& EPDMS$^\dagger$$\uparrow$ \\
\midrule

LeWM
& \multirow{3}{*}{\texttt{navtrain}}
& 82.0 & 67.5 & 81.4 & 98.5 & 66.1 & 79.5 & 79.0 & 66.7 & 13.7
& 39.8
& 66.7 \\

AD-E2E-JEPA
&
& 93.1 & 82.4 & 95.6 & 99.5 & 80.6 & 90.9 & 89.4 & 89.6 & 21.7
& 63.5
& 80.1 \\

\quad + rollout
&
& 94.7 & 80.9 & 93.8 & \textbf{99.7} & 84.0 & 92.5 & 87.9 & 91.8 & 34.6
& 64.9
& 83.1 \\

\midrule

AD-E2E-JEPA
& \multirow{7}{*}{\texttt{trainval}}
& 92.8 & 82.8 & 95.3 & 99.4 & 82.0 & 90.6 & 89.8 & 88.5 & 19.5
& 63.2
& 80.1 \\

\quad + rollout
&
& 95.2 & 82.0 & 94.4 & 99.6 & 85.2 & 93.3 & 89.3 & 92.3 & 35.5
& 67.3
& 84.1 \\

\quad + rollout, 512 traj.
&
& 95.9 & 83.6 & 95.5 & 99.6 & \textbf{85.6} & 94.4 & 90.4 & 93.5 & 36.9
& 69.2
& 85.0 \\

\quad + rollout, 1024 traj.
&
& 96.4 & 84.2 & 96.1 & 99.6 & \textbf{85.6} & 95.0 & 91.0 & 94.1 & 38.5
& 70.5
& 85.5 \\

\quad + rollout, 2048 traj.
&
& 96.7 & 84.8 & 96.4 & \textbf{99.7}
& \textbf{85.6} & 95.4 & 91.5 & 94.5 & 40.9
& 71.5
& 86.0 \\

\quad + rollout, 4096 traj.
&
& 96.7 & 85.4 & 96.4 & \textbf{99.7} 
& 85.5 & 95.6 & 91.5 & 95.0 & 42.5
& 72.1
& 86.3 \\

\quad + rollout, 8192 traj.
&
& \textbf{96.9} & \textbf{86.0} & \textbf{96.7} & \textbf{99.7}
& 85.4 & \textbf{95.7} & \textbf{91.8} & \textbf{95.1} & \textbf{43.8}
& \textbf{72.9}
& \textbf{86.5} \\

\bottomrule
\end{tabular}
\end{table*}

\subsection{Downstream Imitation Learning}
\label{appendix:imitation_learning}

\begin{figure}[ht!]
  \centering
  \includegraphics[width=\linewidth]{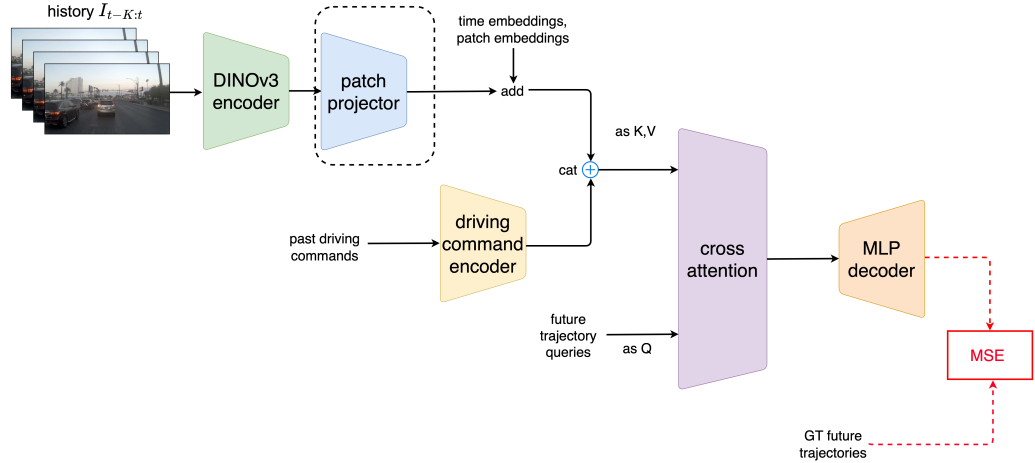}
  \caption{Downstream imitation learning architecture.}
  \label{fig:downstream_imitation_learning_architecture}
\end{figure}

For downstream imitation learning, we adapt the simple ViT-based architecture used in Drive-JEPA~\citep{wang2026drive}. We replace the encoder with a pretrained DINOv3 encoder and append the proposed patch projector. We additionally incorporate temporal embeddings and patch positional embeddings, with the patch positional embeddings shared across frames. The encoded driving command is concatenated with the projected patch embeddings. The future trajectory is encoded as a query and processed by a cross-attention layer, followed by an MLP that predicts the ground-truth future trajectory. We train the network using an MSE loss. We compare the performance of a randomly initialized projector with that of the pretrained projector obtained during AD-E2E-JEPA world modeling. The input context consists of four front-camera frames, each represented as a $3\times256\times512$ tensor. The architecture is illustrated in Figure~\ref{fig:downstream_imitation_learning_architecture}.

\end{document}